\documentclass[10pt]{article}

\usepackage[top=4cm,bottom=3cm,left=2.54cm,right=2.54cm]{geometry}
\usepackage{graphicx}
\usepackage{makecell}
\usepackage{multirow}
\usepackage{mathtools}
\usepackage{xspace}
\usepackage{amsmath}
\usepackage{amssymb}
\usepackage{booktabs}
\usepackage{pifont}
\usepackage{amsthm} 
\usepackage[inline]{enumitem}
\usepackage[table]{xcolor}
 \usepackage{url}

\theoremstyle{plain}
\theoremstyle{definition}
\theoremstyle{remark}

\newcommand{\methodname}{Super-\textsc{DeepG}\xspace}
\newcommand{\deepgname}{\textsc{DeepG}\xspace}

\newcommand{\greencheck}{\textbf{\ding{51}}}
\newcommand{\redcross}{\ding{55}}

\begin{document}

\title{Certified geometric robustness - \methodname}

\author{Noémie Cohen$^{1,2}$, Mélanie Ducoffe$^1$, Timothée Fayard$^2$, Christophe Gabreau$^1$,\\
  Claire Pagetti$^2$, Xavier Pucel$^2$\\
$^1$ Airbus, $^2$ French Aerospace Lab, ONERA}

\maketitle

\begin{center}
\footnotesize
© 2026 IEEE.  Personal use of this material is permitted.  Permission from IEEE must be obtained for all other uses, in any current or future media, including reprinting/republishing this material for advertising or promotional purposes, creating new collective works, for resale or redistribution to servers or lists, or reuse of any copyrighted component of this work in other works.
\end{center}

\begin{abstract}
Safety-critical applications are required to perform as expected in normal operations.
Image processing functions are often required to be insensitive to small geometric perturbations such as rotation, scaling, shearing or translation.
This paper addresses the formal verification of neural networks against geometric perturbations on their image dataset.
Our method \methodname improves the reasoning used in linear relaxation techniques and Lipschitz optimization, and provides an implementation that leverages GPU hardware.
By doing so, \methodname achieves both precision and computational efficiency of robustness certification, to an extent that outperforms prior work. 
\methodname is shared as an open-source tool on GitHub\footnote{\texttt{https://github.com/NoCohen66/superDeepG}}.
\end{abstract}

\section{Introduction}

Neural networks are ubiquitously used across various safety-critical applications, such as Cyber-Physical Systems. We aim to verify neural
networks used for aircraft runway detection during the landing
phase, utilizing the LARD dataset \cite{bougacha2026lard}. For such applications, it is imperative to ensure that the system satisfies a set of safety properties. One particular safety property is the \textit{robustness} of a function, ensuring that perturbations on its inputs do not alter its outputs in an undesirable manner. A reliable approach to check robustness 
is to apply formal methods. 
Practically, a feed-forward neural network can be described as a function $f : \mathbb{R}^n \to \mathbb{R}^m$. Given an input domain $\mathcal{C} \subseteq \mathbb{R}^n$ and a robustness property $\mathcal{P} \subseteq \mathbb{R}^m$, the verification problem can be formulated as:
\begin{equation}
\forall x_0 \in \mathcal{C}, \text{ does } f(x_0) \text{ satisfy } \mathcal{P}?
\label{eq:generalproperty}
\end{equation}
To solve the verification problem in Equation \ref{eq:generalproperty}, two main tasks must be carried out. First, the input domain $\mathcal{C}$ must be tightly defined, so that it accurately reflects the intended perturbations. Second, verifiers, such as Marabou \cite{marabou}, are then used to check the property.
In the literature, the input domain $\mathcal{C}$ is often represented as a norm-based perturbation region around a fixed input $x$.
\emph{White noise} perturbation for instance is defined by the $l_\infty$ norm ball, i.e., $\mathcal{C} = \{x_0 \in \mathbb{R}^n \mid \|x - x_0\|_\infty \leq \epsilon\}$. This type of domain is not specific to any class of function implemented by the neural network.

In this work, we focus on image processing functions, and investigate image perturbations induced by geometric transformations. Unlike norm-based perturbations, which alter pixel intensities directly while keeping their positions fixed, these transformations shift and interpolate pixel values, complicating the  definition of the input domain $\mathcal{C}$.
This paper addresses the problem of efficiently determining a precise geometry-based input domain.

Several methods exist for this purpose. 
However, none of them can be described as both:
\begin{enumerate*}[label=(\arabic*)]
    \item precise, ensuring that the input domain is tight enough for effective property verification;
    \item scalable, allowing to verify large neural networks;
    \item fast, leveraging GPU utilization for enhanced performance.
\end{enumerate*}
We build on the \deepgname~\cite{deepG} approach due to its strong theoretical foundation and precision in geometry-based input domain verification, despite its computational cost.
Our main contributions are as follows:
\begin{enumerate} 
\item We reformulate the mathematical foundations underlying the different optimization steps employed by \deepgname. 
We eliminate the use of iterative methods, propose new formulations for each optimization problem and achieve equivalent precision. 
\item This change not only removes a major computational bottleneck but also enables efficient GPU parallelization, yielding up to a 526× speedup. We achieve significant geometric certification results on TinyImageNet without training specifically for geometric robustness.
\item We provide an open-source implementation within the \texttt{auto$\_$LiRPA}~\cite{xu2020automatic} verifier, encapsulating the geometric transformations in a \texttt{CustomOp}, which simplifies integration and usability for practitioners. 
\end{enumerate}
Section~\ref{sec:background} recalls the mathematical background and the problem statement. Section~\ref{section:related_work} provides an overview of related work in the field. Section~\ref{sec:deepg} summarizes the \deepgname approach and its use of computationally expensive iterative methods to solve mathematical problems. Section~\ref{sec:fromdeepgtosuperdeepg} 
presents \methodname and our efficient solutions to these mathematical problems.
Section~\ref{sec:expe} shows experiments to compare our method with state-of-the-art approaches on benchmark networks. Section~\ref{sec:conclusion} concludes the paper.

\section{Background}\label{sec:background}

We consider images composed of $c$ channels of matrices $n\times m$ where $n, m$ are odd integers. 
Since all computations and perturbations are applied channel-wise, we restrict the presentation to a single channel to simplify notations. In particular, images become 2D tensors.
Let $A, B \in \mathbb{R}^{n\times m}$, $A \geq B$ (resp. $A \leq B$) denotes $\forall i,j$, $A_{i,j} \geq B_{i,j}$ (resp. $A_{i,j} \leq B_{i,j}$). We use $\underline{A}$ and $\overline{A}$ to denote the lower and upper bounds of a matrix \( A \), i.e., $\underline{A} \leq A \leq \overline{A}$.
\subsection{Geometric transformation}
A geometric image transformation is usually defined as a function $g : \mathbb{R}^d \times \mathbb{R}^{n\times m} \to \mathbb{R}^{n\times m}$, where $\kappa \in \mathbb{R}^d$ is the transformation parameter (e.g. rotation angle). In robustness verification, each image is analyzed independently, thus we consider a single image $x \in \mathbb{R}^{n\times m}$, and study its geometric transformation $g_x : \mathbb{R}^d \to \mathbb{R}^{n\times m}$.
We denote $p_{i,j}(\kappa)$ the pixel value at position $(i, j)$ in the transformed image $g_x(\kappa)$. The value associated to this pixel is defined by:
\begin{equation}
{p}_{i,j}(\kappa) = I_x \circ T_\kappa^{-1}(i, j)
\label{eq:interpolation_rotation_composition}
\end{equation}
Where $I_x$ is an interpolation function, and $T_\kappa$ the geometric transform. $T_\kappa$ may correspond to a rotation, a translation, a shearing, or a scaling, all of which have a unique inverse $T_\kappa^{-1}$. For simplicity and clarity, we provide detailed explanations for rotation, but the formulas for the other transformations can be found in Appendix \ref{appendix:geom_transfo}. Transformation compositions are not detailed but fall in the same mathematical framework.

Let $\kappa$ be a rotation angle $\theta \in \mathbb{R}$, the inverse transform $T_\kappa^{-1}: \mathbb{N}^2 \to \mathbb{R}^2$ maps the coordinates of the transformed pixel back to the original image’s coordinates. It is represented by the matrix:
\begin{equation}
        R^{-1}_{\theta} (i,j) = \begin{pmatrix} \cos \theta & \sin \theta \\ -\sin \theta  &  \cos \theta \end{pmatrix} \begin{pmatrix} i-\frac{n}{2} \\ j - \frac{m}{2} \end{pmatrix}
        + \begin{pmatrix} \frac{n}{2} \\ \frac{m}{2} \end{pmatrix}
\label{eq:rotation}
\end{equation}
The interpolation $I_x: \mathbb{R}^2 \to \mathbb{R}$ takes as input non-integer pixels coordinates resulting from the rotation and outputs an interpolated pixel value. Among the many existing interpolation functions, we consider the most common \cite{deepG, CGT, batten}, i.e. the standard bilinear interpolation:
 \begin{equation}
\begin{split}
    I_x(i, j) &= \sum_{k=1}^{n} \sum_{l=1}^{m} p_{k,l}(0) \cdot \max(0,1-|i - k|) \\
    &\quad \quad \quad \quad \quad \; \cdot \max(0,1-|j-l|)
\end{split}
\label{eq:interpolation}
\end{equation}
\subsection{Geometric Robustness Verification}
Geometric robustness, derived from Equation \eqref{eq:generalproperty}, is to verify whether a neural network $f$ satisfies the robustness property $\mathcal{P}$ for all transformed images $g_x(\theta)$:
\begin{equation}
\forall \theta \in [\underline{\theta}, \overline{\theta}],  \quad \text{ does } f(g_x(\theta)) \text{ satisfy } \mathcal{P}?
\label{eq:gpropertygeom}
\end{equation}
Due to the highly non-linear nature of geometric transformations, computing the exact set $g_x([\underline{\theta}, \overline{\theta}])$ is not tractable. To address this, we apply abstract interpretation~\cite{cousot,ai2} and define an \textit{abstract domain} $\mathcal{C}$ that over-approximates $g_x([\underline{\theta},\overline{\theta}])$.
State-of-the art methods~\cite{deepG, semantifyNN, CGT} compute linear bounds
$\underline{A}, \underline{B}, \overline{A},\overline{B} \in  \mathbb{R}^{n\times m}$ such that:
$
\underline{A} \cdot \theta + \underline{B} \leq  g_x(\theta) \leq  \overline{A} \cdot \theta + \overline{B}$, and define the abstract domain as:
\begin{equation}
\mathcal{C} = \left\{x_0 \mid \underline{A} \cdot \theta + \underline{B} \leq  x_0 \leq  \overline{A} \cdot \theta + \overline{B} \right\}
\label{eq:linear_eq_pixel}
\end{equation}
Thus, instead of Equation~\eqref{eq:gpropertygeom}, we address the problem:
\begin{equation}
\forall x_0 \in \mathcal{C}  \supseteq  g_x([\underline{\theta}, \overline{\theta}]) , \text{ does } f(x_0) \text{ satisfies } \mathcal{P}? 
\label{eq:gpropertygeomapprox}
\end{equation}
Interval Bound Propagation (IBP) can be seen as the special case where $\underline{A} = \overline{A} = 0$ in Equation~\eqref{eq:linear_eq_pixel}.

Finally, in this paper we focus on image classification. The function $f : \mathbb{R}^{n \times m} \to \mathbb{R}^k$ denotes the classifier that computes the score $f_k$ associated to each of the $k$ classes.
For the input image $x$, with ground truth class $c$, we assume $f$ is correct, 
meaning that $f_c(x) > f_i(x)$ for all $i\not=c$.
Abstract interpretation techniques can be extended to the function $f$, yielding linear bounds $\underline{f(x_0)}$ and $\overline{f(x_0)}$ for each class.
The standard classification robustness property $\mathcal{P}$ 
is expressed as:
\begin{equation}
\mathcal{P} \equiv 
\min_{i \neq c} \left\{ \underline{f_c(x_0)} - \overline{f_i(x_0)} \right\} > 0
\label{eq:verified-property}
\end{equation}

\section{Related Work}\label{section:related_work}

Formal verification methods such as Satisfiability Modulo Theory, MILP, and abstract interpretation are widely used for verifying neural network robustness \cite{survey_methods_liu, urban_survey, survey_methods_Meng}. 
Among these, 
abstract interpretation is particularly suitable for scaling up to large networks with high-dimensional inputs. 
The choice of the abstract domain impacts the precision and the computational efficiency.
Tighter domains yield more precise abstractions at the price of increased computational cost.
Linear relaxation based perturbation analysis (LiRPA), used in CROWN~\cite{CROWN} and DeepPoly~\cite{deepoly},
propagates \emph{linear} bound-based abstract domains (see Equation~\ref{eq:linear_eq_pixel}), improving precision with respect to IBP. 

FGV \cite{CGT} and semantify-NN \cite{semantifyNN}, have attempted to linearize geometric transformations by over-approximating each operator involved (e.g., \(\cos\), \(\sin\), $\max$, $\operatorname{abs}$ in Equations~(\ref{eq:rotation},\ref{eq:interpolation})) to fit within existing verification frameworks. 
However, these over-approximations often lead to loose bounds. This imprecision is even worse when relying on less precise techniques like IBP, as done by FGV and semantify-NN. Consequently, these methods sacrifice precision for scalability and improved runtime performance.  

\deepgname~\cite{deepG} considers the geometric transformation as a single function when producing linear bounds. This approach yields very precise bounds upon which robustness verification can be built. However, its computational cost is high due to its reliance on a linear programming solver and its design, which is not optimized for GPUs.
PWL \cite{batten} builds upon \deepgname by defining the domain $\mathcal{C}$ as a convex polytope, and uses a MILP solver to verify the property, which has a hard time scaling up to large neural networks.

\begin{table}
\centering
\setlength{\tabcolsep}{5pt}
\begin{tabular}{|*{5}{c|}}
\hline
$g_x(\cdot)$ & \deepgname & PWL & FGV & \methodname \\
\hline
Precise   & \greencheck & \greencheck\greencheck & \redcross  & \greencheck \\
\hline
Fast                    &   \redcross  & \redcross &    \greencheck\greencheck      &  \greencheck \\
\hline
\end{tabular}
\caption{Tools implementing geometric abstract domains}
\label{tab:comparison_of_approaches}
\end{table} 
\setlength{\intextsep}{0pt}  
\setlength{\textfloatsep}{0pt} 
From our review and experiments summarized in Table~\ref{tab:comparison_of_approaches}, we draw two conclusions.
\begin{enumerate}[noitemsep, topsep=0pt]
    \item The mathematical foundation of \deepgname is the key to precise abstractions. Its efficiency can be improved without impacting its precision.
    \item All approaches have a hyperparameter for splitting the interval $[\underline{\theta}, \overline{\theta}]$ and making the verification per sub-interval. Increasing the number of sub-intervals is a very effective way to gain precision. FGV and Semantify-NN leverage their speed by relying heavily on this mechanism. In contrast, \deepgname and PWL are relatively slow and refrain from splitting too much.
\end{enumerate}
These statements motivate our approach called \methodname, that improves the mathematical foundation of \deepgname and uses solutions that can be efficiently implemented on GPU architectures. Furthermore, in order to demonstrate that \methodname outperforms other tools, it is imperative to be cautious in the choice of hyperparameters to produce fair and meaningful comparisons.

Other techniques have been investigated for geometric robustness, such as randomized smoothing \cite{fischer2021scalable}, \cite{li2021tss} or unsound methods \cite{zhang2025scalable}. However, they fail to provide the deterministic and sound guarantees we require.

\section{In-depth \deepgname analysis}\label{sec:deepg}

\deepgname computes 
the abstract domain~$\mathcal{C}$ with linear bounds as in Equation~\eqref{eq:linear_eq_pixel}.
It takes as input an image $x$, an interval of rotation angles $[\underline{\theta}, \overline{\theta}]$, 
and it is tuned by two hyper-parameters P and $\varepsilon$. It produces as output the four matrices $\underline{A}$, $\underline{B}$, $\overline{A}$ and $\overline{B}$ that define $\mathcal{C}$. Then, it uses the solver ERAN \cite{eran} to produce the linear bounds $\underline{f}$ and $\overline{f}$ and verify property~\eqref{eq:verified-property}.
\deepgname{} decomposes the problem into three main steps.  
First, it computes initial bounds that hold for some sampled points, but may not generalize to the full domain.  
Second, it (over)estimates the Lipschitz constant for the transform.  
Finally, it derives sound bounds by applying a margin on the initial bounds. This margin is obtained by sampling and using Lipschitz certificates.
\subsection*{Step 1: Unsound Linear Bounds using LP}
\begin{figure}[hbt]
    \centering
    \includegraphics[width=0.55\linewidth]{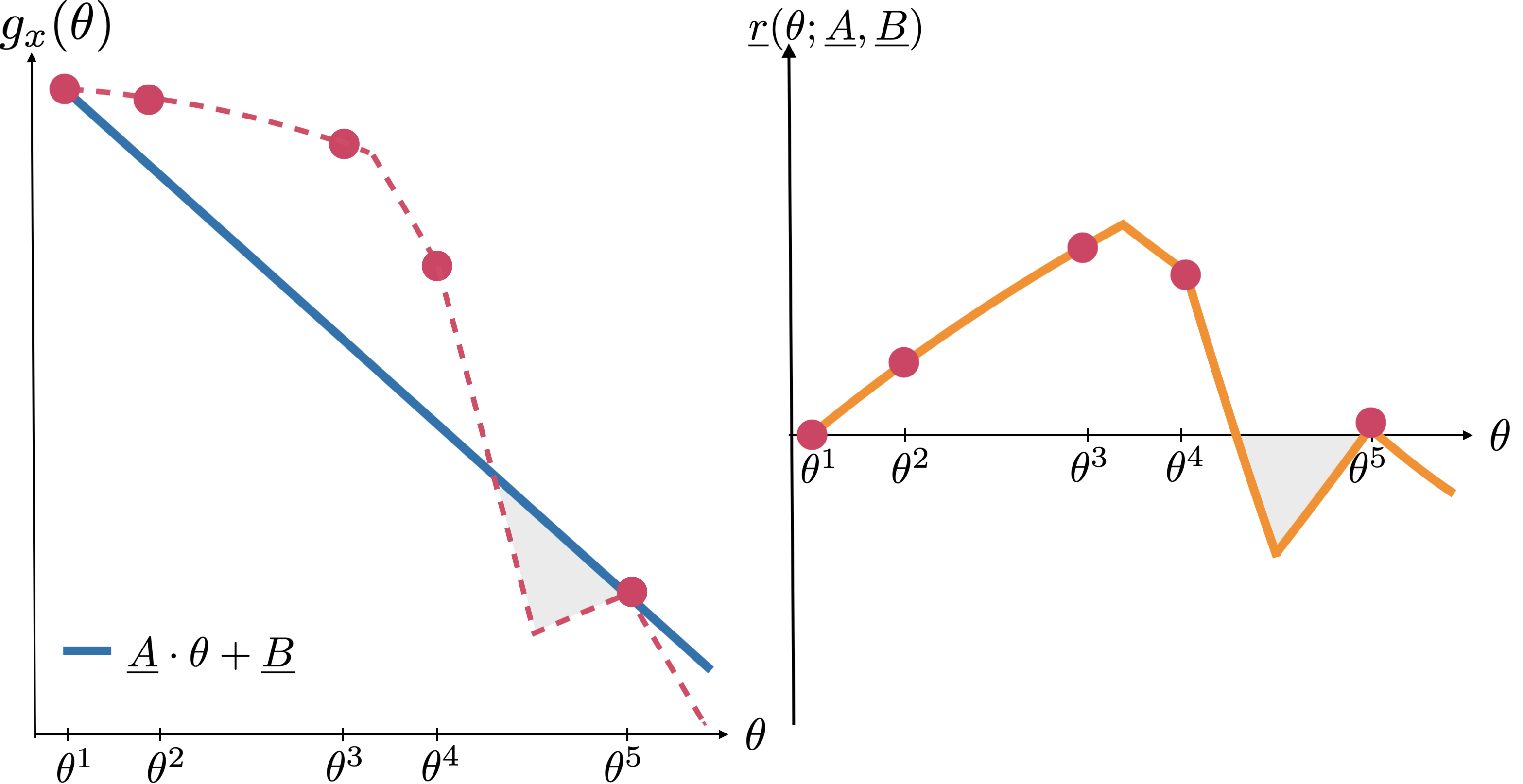}
    \caption{Initial lower linear bound for the sampled values of one pixel at varying transformation values. Left: the pixel value (dashed), sampled points (red) and a lower linear bound (blue) for the samples. Right: the residual function $\underline{r}(\theta^p; \underline{A}, \underline{B})$, representing the difference between the real pixel value $g_x(\theta)$ and the initial lower bound.}
    \label{fig:step1}
\end{figure}

\deepgname\ first samples \(P\) points \(\theta^p\) for \(p=1,\dots,P\) within the input domain \([ \underline{\theta}, \overline{\theta} ]\)
and computes the associated transformed images \( g_x(\theta^p) \).
It computes 
a value of $\underline{A}, \underline{B}, \overline{A}, \overline{B}$
with those empirical data  as shown in Figure \ref{fig:step1}. 
To do so, we define the lower residual \(\underline{r}(\theta^p; \underline{A}, \underline{B}) = g_x(\theta^p) - \underline{A} \cdot \theta^p - \underline{B}\).
 $\underline{A}, \underline{B}$ are the solutions of an optimization problem that minimizes 
 the average residual while enforcing non-negativity for every sampled point.
 This is formulated as the following \(\mathfrak{P}\) optimization problem:
\begin{equation}
\begin{aligned}
\min_{\underline{A}, \underline{B}} \quad & \left\{ \frac{1}{P} \sum_{p=1}^{P} \underline{r}(\theta^p; \underline{A}, \underline{B}) \right\} \\
\text{subject to} \quad & \underline{r}(\theta^p; \underline{A}, \underline{B}) \geq 0, \quad \forall p = 1,\ldots, P
\label{eq:LP}
\end{aligned}    
\end{equation}
Here, $\underline{A}$, $\underline{B}$, and the residual $\underline{r}$ are matrices. The $\min$ operation is therefore interpreted element-wise.
In practice, \deepgname\ solves Equation~\eqref{eq:LP} as a linear programming problem 
with the Gurobi solver. A similar process is applied to compute the upper bound. This step requires solving independent scalar optimization problems for each pixel and is computationally expensive.
\subsection*{Step 2: Computing a Lipschitz constant}
\deepgname computes a valid Lipschitz constant $\mathcal{L}$ for the residual function, that provides an upper bound to how much the residual can vary between two sampled points. 
The tightest Lipschitz constant $\mathcal{L}^{*}$ is defined as the supremum of the gradient norm of the residual function:
\[
\mathcal{L}^{*}
  = \sup_{\theta \in \left[\underline{\theta}, \overline{\theta}\right]}
    \left\|
      \frac{\partial \underline{r}(\theta)}{\partial \theta}
    \right\|_{\infty}
\]
\deepgname computes an approximate upper-bound $\mathcal{L} \ge \mathcal{L}^{*}$.
In addition to the constant $\mathcal{L}$ valid on all $\left[\underline{\theta}, \overline{\theta}\right]$, \deepgname can compute a Lipschitz constant $\mathcal{L}(\theta_1, \theta_2)$ for any sub-interval $\left[\theta_1, \theta_2\right]$.
\subsection*{Step 3: Correction of the approximation error}
While the bound $\underline{A}\cdot\theta+\underline{B}$ computed in Step 1 is a lower bound for the sampled points, it may fail to be a lower bound for the whole interval $[\underline{\theta}, \overline{\theta}]$, as illustrated in grey in Figure~\ref{fig:step1}.
The goal of this step is to add a correction term $\underline \delta \leq 0$, to obtain a valid lower bound over the entire domain:
\begin{equation*}
    \forall \theta \in [\underline{\theta}, \overline{\theta}], \;g_x(\theta) \geq \underline{A}\,\theta + \underline{B} + \underline{\delta}
\end{equation*}
The minimum value of $\delta$ that ensures a correct lower bound is given by $\underline{\delta}^* =\min_{\theta \in [\underline{\theta}, \overline{\theta}]} \underline{r}(\theta, \underline{A}, \underline{B})$. However, computing $\delta^*$ is computationally hard due to the nonlinearity and nonconvexity of the function. Instead, \deepgname computes an approximate term $\underline{\delta} \leq \underline{\delta}^*$ by applying a Lipschitz certificate. From the Liptschitz certificate property:
\begin{equation}\label{equ:lipschitz_certificate}
    \min_{\theta \in [\underline{\theta}, \overline{\theta}]} \underline{r}(\theta; \underline{A}, \underline{B}) \geq \underline{r} \left( \frac{\underline{\theta} + \overline{\theta}}{2}; \underline{A}, \underline{B} \right) - \mathcal{L} \left| \frac{\overline{\theta} - \underline{\theta}}{2} \right|
\end{equation}
Thus, the right-hand side of Equation~\eqref{equ:lipschitz_certificate} is a valid value for our correction term $\underline{\delta}$. Splitting the domain $[\underline{\theta}, \overline{\theta}]$ is a natural way to obtain tighter correction terms. \deepgname partitions the input range into M intervals \([ \underline{\theta}, \overline{\theta} ] = \bigcup_{i=0 \dots M} [\theta_i, \theta_{i+1}]\), and searches for a minimum in each partition:
\begin{displaymath}
 \min_{\theta \in [\underline{\theta}, \overline{\theta}]} \underline{r}(\theta; \underline{A}, \underline{B}) = \min_{i=0 \dots M} \min_{\theta \in [\theta_i, \theta_{i+1}]} \underline{r}(\theta; \underline{A}, \underline{B})
\end{displaymath}
Thus, applying the Lipschitz certificate to each partition yields the following:
\begin{equation}\label{equ:doo_step}
\begin{aligned}
    \min_{\theta \in [\underline{\theta}, \overline{\theta}]} \underline{r}(\theta; \underline{A}, \underline{B}) \geq& \min_{i = 0,\ldots,M} \underline{r} \left( \frac{\theta_i + \theta_{i+1}}{2}; \underline{A}, \underline{B} \right) \\
    &- \mathcal{L}(\theta_i, \theta_{i+1}) \left| \frac{\theta_{i+1} - \theta_i}{2} \right|
\end{aligned}
\end{equation}
The partition \(j\), that contains the minimum, is further split, leading to:
\(
[\underline{\theta}, \overline{\theta}] =  \bigcup_{i=0 \dots j-1, j+1 \dots M} [\theta_i, \theta_{i+1}] 
 \cup \left[\theta_j, \frac{\theta_j + \theta_{j+1}}{2}\right] 
 \cup \left[\frac{\theta_j + \theta_{j+1}}{2}, \theta_{j+1}\right]
\). The iteration step is repeated
as many times as needed. Doing so,
\deepgname revisits the well-known DOO algorithm \cite{DOO}
which guarantees convergence to the exact minimum, given sufficient iterations, and ensures that the bounds remain valid throughout the optimization process.
\begin{figure}[hbt]
    \centering
    \includegraphics[width=0.55\linewidth]{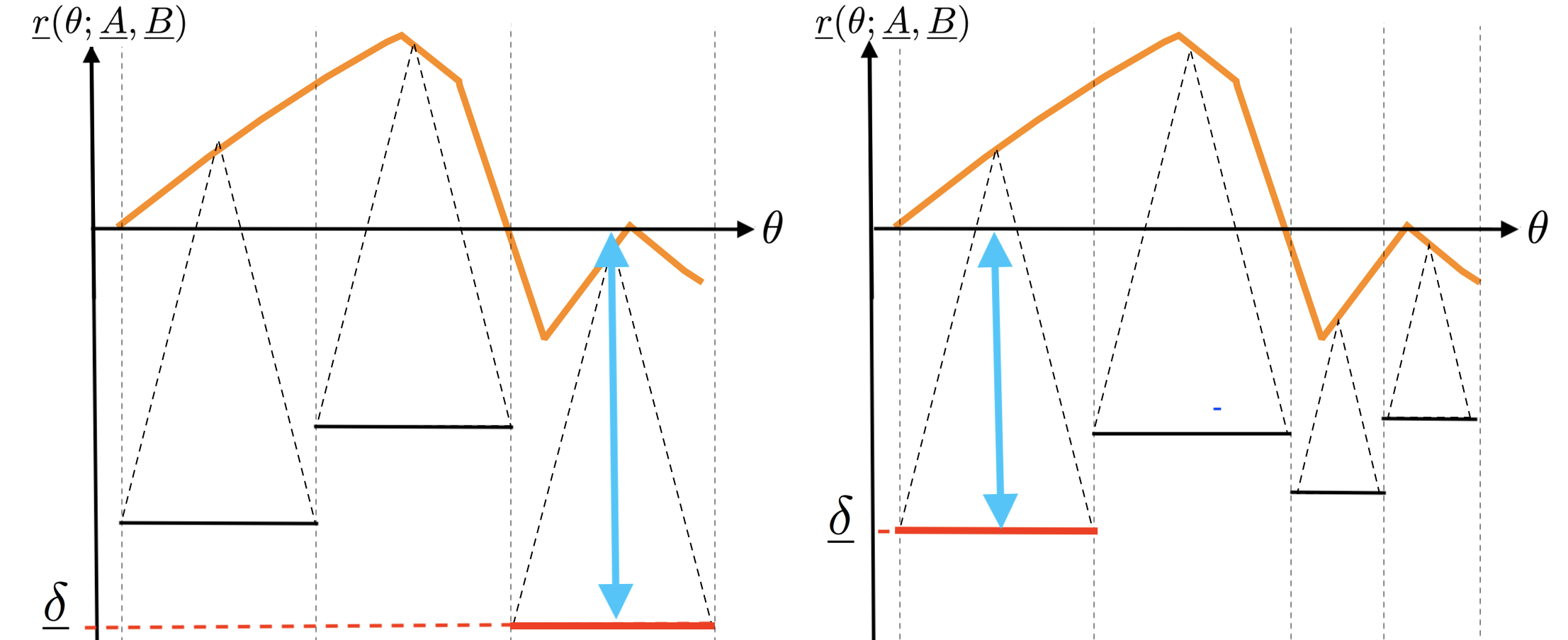}
    \caption{Illustration of the DOO algorithm used in \deepgname. Left: Residual function with an initial partition (\(M=3\)) and penalty-adjusted bounds. Right: Partition update by splitting the interval with the maximum penalty (in red).}
    \label{fig:step2}
\end{figure}
The procedure terminates when the size of the sub-interval associated with the current minimum certificate satisfies:
\begin{equation}
    \left| \theta_{i+1} - \theta_i \right| < 2\varepsilon/\mathcal{L}(\theta_i, \theta_i+1)
\label{eq:relationepsilon}
\end{equation}
where $\varepsilon$ is an input hyperparameter.
Let \(\underline{\delta}\) denote the resulting certified lower bound. A similar process is applied on the upper bound. The corrected bounds from Step~1 then satisfy:
\begin{equation}
    \underline{A} \cdot \theta + \underline{B} + \underline{\delta} \leq g_x(\theta) \leq \overline{A} \cdot \theta + \overline{B} + \overline{\delta}
\label{eq:step3deepg}
\end{equation}
A major limitation of Step 3 is the iterative nature of the DOO implementation. As a result, the direct application of this approach in modern verification frameworks is not feasible as is.

\section{From \deepgname to \methodname}\label{sec:fromdeepgtosuperdeepg}

\methodname improves on \deepgname in two ways. It removes iterative algorithms and replaces them with static \emph{meshed} and \emph{tensorized} formula compatible with GPU implementation. 
\methodname has the same inputs $x$ and $[\underline{\theta}, \overline{\theta}]$, and the same outputs $\underline{A}$, $\underline{B}$, $\overline{A}$, $\overline{B}$ as \deepgname. It has the hyperparameter $P$ for step 1 and $N$ for step 3. 

\subsection{Step 1-bis: Analytical Unsound Bounds} 
This step also computes initial unsound linear bounds (with $\underline{A}$, $\underline{B}$, $\overline{A}$, $\overline{B}$). 
Like \deepgname, we sample \(P\) points \(\theta^p\) for \(p = 1,\dots,P\) within the input domain \([ \underline{\theta}, \overline{\theta} ]\)
and compute the associated images \( g_x(\theta^p) \). However, we solve the 
optimization problem  \(\mathfrak{P}\) of Equation~\eqref{eq:LP} differently.

We have demonstrated in Appendix \ref{proofbycontradiction} that any solution \(\underline{A}, \underline{B}\)
of \(\mathfrak{P}\) satisfies \(\exists \theta^p, \underline{A}\theta^p+ \underline{B} = g_x(\theta^p)\).
Thus, adding the constraint $\underline{r}(\theta^1,\underline{A},\underline{B}) = 0 \vee \ldots \vee \underline{r}(\theta^P,\underline{A},\underline{B}) = 0$ to \(\mathfrak{P}\) transforms it into an equivalent problem.
Solving this transformed problem is equivalent 
to solve $P$ sub-problems \(\mathfrak{P}^p\),
each with the constraint $\underline{r}(\theta^p,\underline{A},\underline{B}) = 0$ and 
take the optimal solution among the $P$ sub-solutions.
Let us denote by $\underline{A}^p, \underline{B}^p$
the solutions of \(\mathfrak{P}^p\).
Since they satisfy the constraint
$\underline{r}(\theta^p,\underline{A^p},\underline{B^p}) = 0$, 
there is a relation between them as
\(\underline{B}^p = g_x(\theta^p) - \underline{A}\theta^p\). 
The problem  \(\mathfrak{P}^p\) thus only
optimizes $\underline{A^p}$, 
as $\underline{B^p}$ can be derived.
The other constraint and the optimization criteria 
of \(\mathfrak{P}^p\) 
is on the residual,  \(\underline{r}(\theta; \underline{A^p}, \underline{B^p}) = g_x(\theta) - \underline{A^p} \cdot \theta - \underline{B^p} = g_x(\theta) - \underline{A^p} \cdot \theta - g_x(\theta^p) + \underline{A^p}\theta^p= (g_x(\theta) - g_x(\theta^p))-  \underline{A^p} \cdot (\theta - \theta^p)\).
We denote \(
\underline{r_p}(\theta; \underline{A^p})  = \underline{r}(\theta; \underline{A^p}, \underline{B^p})\).
The sub-problem \(\mathfrak{P}^p\) is thus defined by:
\begin{equation}
\label{eq:LPnewcrit}
\begin{aligned}
\min_{\underline{A^p}}\quad & \left\{  \frac{1}{P}\sum_{k=1}^{P}
\underline{r}_p(\theta^k; \underline{A^p})
\right\}\\
\text{subject to: } \quad &
\underline{r}_p(\theta^k; \underline{A^p}) \geq 0
\quad \text{for } k = 1, \ldots, P
\end{aligned}
\end{equation}

To solve \(\mathfrak{P}^p\),
we consider the two sets $\mathcal{K}^+ = \{ k \in 1, \dots, P \mid \theta^k - \theta^p > 0 \}$ and $\mathcal{K}^- = \{ k \in 1, \dots, P  \mid \theta^k - \theta^p < 0 \}$. In order to satisfy constraint $\underline{r}_p(\theta^k; \underline{A^p}) \geq 0$, \(\underline{A^p}\) must lie within the interval:
\begin{equation}
\underline{A^p} \in
   \left[\max_{k \in \mathcal{K}^-}\!\frac{g_x(\theta^p) - g_x(\theta^k)}{\theta^p - \theta^k},
   \;\,
   \min_{k \in \mathcal{K}^+}\!\frac{g_x(\theta^p) - g_x(\theta^k)}{\theta^p - \theta^k}\right]
\label{eq:step_first_interval}
\end{equation}
Hence, an optimal \(\underline{A^p}\) is given by:
\begin{equation}
\underline{A^p} \;=\;
\begin{cases}
\displaystyle
   \min\limits_{k \in \mathcal{K}^+}\!\frac{g_x(\theta^p) - g_x(\theta^k)}{\theta^p - \theta^k}
   & \text{if } 
   \Bigl(\,\theta^p - \sum\limits_{k=1}^{P} \frac{\theta^k}{P}\Bigr) \geq 0,
   \\
\displaystyle
   \max\limits_{k \in \mathcal{K}^-}\!\frac{g_x(\theta^p) - g_x(\theta^k)}{\theta^p - \theta^k}
   & \text{otherwise}.
\end{cases}
\label{eq:step1bis-ap}
\end{equation}
The interval notation in Equation \eqref{eq:step_first_interval}, as well as the $\max$ and $\min$ operations in Equation \eqref{eq:step_first_interval} and Equation \eqref{eq:step1bis-ap}, are to be interpreted element-wise.
The optimal solution $\underline{A}, \underline{B}$ of \(\mathfrak{P}\) is
$\underline{A^{p_{opt}}}, \underline{B^{p_{opt}}}$ such that
$\underline{A^{p_{opt}}}, \underline{B^{p_{opt}}}$ is a solution of  \(\mathfrak{P}^{p_{opt}}\) 
and
$\frac{1}{P}\sum_{k=1}^{P} \underline{r}_p(\theta^k; \underline{A^{p_{opt}}})=\min_{p=1,\ldots, P} \frac{1}{P}\sum_{k=1}^{P} \underline{r}_p(\theta^k; \underline{A^{p}})$.

Informally, \methodname enumerates all cases in which the linear relaxation passes through at least one sample point (i.e., $r(\theta^p;A,B) = 0$).  By solving all such subproblems and selecting the best solution, \methodname effectively covers every candidate for an optimal solution in \deepgname's Step~1.  Consequently, both methods yield the same pair $(A^*,B^*)$, ensuring that Step~1-bis of \methodname is indeed equivalent to Step~1 of \deepgname.
This reasoning applies for any $\kappa \in \mathbb{R}$, corresponding to a one-dimensional transformation such as rotation, scaling, and shearing. For the more general case $\kappa \in \mathbb{R}^d$ (e.g., $d=2$ for a translation), detailed formulations are provided in Appendix~\ref{sec:multidunsound}.

The main contribution of our approach is that it supports GPU parallelization, which makes it possible to treat all pixels in parallel. In our experiments, we show empirically that the choice of bounds does not significantly depend on the number of samples: even with a small number of samples, the selected bounds are nearly identical to those chosen with more samples. As in \deepgname, the bounds computed in this step are unsound, and we need to apply a Lipschitz-based margin to ensure they provide a true over-approximation.

\subsection{Step 2-bis: Computing a Lipschitz constant}
For this step, we reuse the efficient PyTorch implementation from PWL to obtain an upper bound $\mathcal{L}(\theta_1, \theta_2)$ for the Lipschitz constant $\mathcal{L}^{*}(\theta_1, \theta_2)$ over any sub-interval $[\theta_1, \theta_2]$.

\subsection{Step 3-bis: Correction of the approximation error}\label{step-3-bis}
\begin{figure}[hbt]
    \centering
    \includegraphics[width=0.55\linewidth]{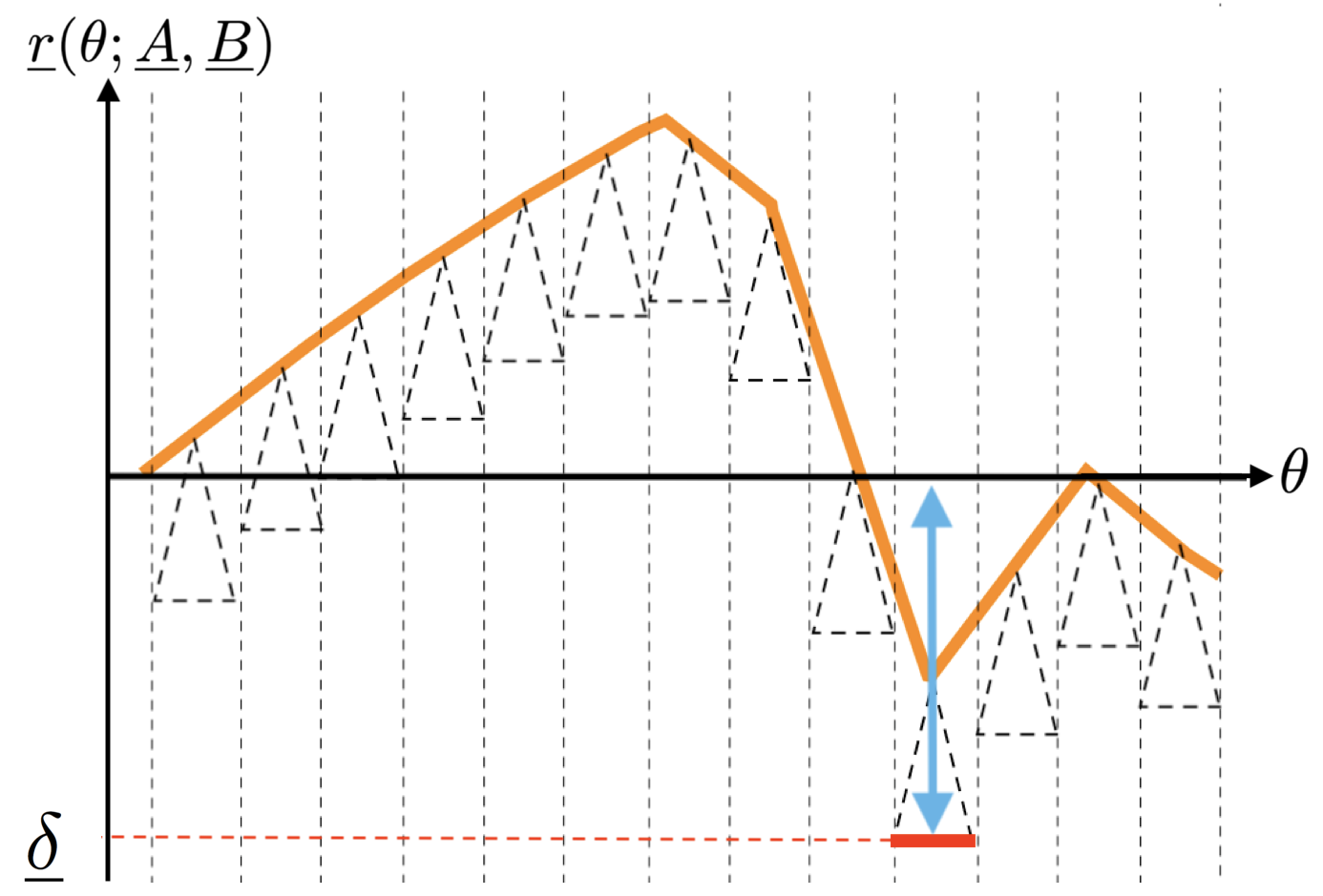}
    \caption{\methodname approach with \(N=13\) partitions. The correction term is in red.}
    \label{fig:step3super}
\end{figure}

As \deepgname, Step 3-bis aims at computing $\underline{\delta}$ and $\overline{\delta}$ to correct  $\underline{A}$, $\underline{B}$ of Step 1-bis.
Instead of applying an iterative strategy, we partition $[\underline{\theta}, \overline{\theta}]$ in $N$ sub-intervals $[\theta_i, \theta_{i+1}]$. Given a bound $\mathcal{L}$ for the Lipschitz constant over $[\underline{\theta}, \overline{\theta}]$, a subdivision number of $N$ entails, by equation \eqref{equ:lipschitz_certificate},
that the correction term is $\underline{\delta}= \min_i \underline{r}(\frac{\theta_i + \theta_{i+1}}{2}, \underline{A}, \underline{B}) - \mathcal{L}|\overline{\theta} - \underline{\theta}| / 2N$. We obtain it in one pass as illustrated in Figure~\ref{fig:step3super}.

All subdivisions are chosen with uniform width: $\left| \theta_{i+1} - \theta_i \right| = \left| \overline{\theta} - \underline{\theta} \right| / N$. To ensure that the correction term is at least as precise as that the one computed by \deepgname, the number of subdivisions \(N\) must be selected based on Equation~\eqref{eq:linkepsilonN} and the hyperparameter $\varepsilon$ as follows:
\begin{equation}
    N \geq \frac{\mathcal L \left| \overline \theta- \underline \theta \right| }{2\varepsilon}
\label{eq:linkepsilonN}
\end{equation}
In practice, we further tighten the correction term $\underline{\delta}$ by computing a bound for the Lipschitz constant for each sub-interval. This is efficiently done on GPUs.
Inside the subdivisions, the $\underline{\delta}$ and $\overline{\delta}$ correction terms are computed and used the same way as in \deepgname in Equations~\eqref{equ:lipschitz_certificate} and \eqref{eq:step3deepg}.
The extension of Step~3-bis to multi-dimensional inputs is detailed in Appendix~\ref{appendix:multi-D}.

\section{Experimental Evaluation}\label{sec:expe}

\begin{table*}[hbt]
\centering
\begin{tabular}{ccccccc}
\toprule
\textbf{\makecell[c]{Dataset/Clean acc. \\ Transformation \\
Config. number}}
& \textbf{Range} 
& \textbf{\makecell{Interval \\ size}}  
& \textbf{\makecell{Certification \\ method}}
& \textbf{Certified (\%)}
& \textbf{\makecell[c]{Time \\   (s/image) }}
\\ \midrule

\multirow{4}{*}{\makecell[c]{MNIST  98.0\% \\ Rotation\\ n°1}} 
& \multirow{2}{*}{\([-30^\circ,30^\circ]\)}
& \multirow{2}{*}{$1^\circ$}
  & \deepgname  & 98.0 & 210.72  \\
&&& \methodname   & \textbf{98.0} & 0.40  \\
\cmidrule(lr){2-7}

& \multirow{3}{*}{\([-30^\circ,30^\circ]\)}
& \multirow{3}{*}{$6^\circ$}
& \deepgname   & 86.0  & 48.38  \\
&&& \methodname    & \textbf{86.0}  & 1.14  \\
&&& \methodname$^*$    & \textbf{86.0}  & 0.13 \\
\midrule
\multirow{3}{*}{\makecell[c]{CIFAR10  74.0\% \\ Rotation n°2}} 
 & \multirow{3}{*}{\([-10^\circ,10^\circ]\)} & \multirow{3}{*}{$1^\circ$}
 & \deepgname   & 65.0  & 2389.70  \\
 &  &  & \methodname    & \textbf{65.0} & 7.58  \\
  &  &  & \methodname$^*$    & \textbf{65.0} & 1.90 \\
\midrule
\multirow{3}{*}{\makecell{MNIST  100.0\% \\ Scaling n°3}} 
& \multirow{3}{*}{\([-2,2]\)}
& \multirow{3}{*}{$4$}
  & \deepgname  & 98.0 & 4.25 & \\
&&& \methodname   & \textbf{99.0} & 0.16  &\\
&&& \methodname$^{*}$   & \textbf{99.0} & 0.063  &\\
\midrule
\multirow{3}{*}{\makecell{CIFAR10  79.0\% \\ Scaling n°4}} 
& \multirow{3}{*}{\([-1,1]\)}
& \multirow{3}{*}{$1$}
  & \deepgname  & 69.0 & 248.13 &\\
&&& \methodname   & \textbf{75.0} & 0.83   & \\
&&& \methodname$^{*}$   & \textbf{75.0} & 0.20   & \\
\midrule
\multirow{3}{*}{\makecell{MNIST  100.0\% \\ Shearing n°5}} 
& \multirow{3}{*}{\([-2,2]\)}
& \multirow{3}{*}{$4$}
  & \deepgname  & 98.0  & 4.30 &\\
&&& \methodname  & \textbf{100.0}  & 0.16   &\\
&&& \methodname$^{*}$  & \textbf{100.0}  & 0.065   &\\
\midrule
\multirow{3}{*}{\makecell{CIFAR10  72.0\% \\ Shearing n°6}} 
& \multirow{3}{*}{\([-2,2]\)}
& \multirow{3}{*}{$2$}
  & \deepgname  & 69.0 & 278.29 &\\
&&& \methodname    & \textbf{69.0} & 0.71    &\\
&&& \methodname$^*$    & \textbf{69.0} & 0.18    &\\
\midrule
\multirow{2}{*}{\makecell{MNIST  99.0\% \\ Translation n°7}} 
& \multirow{2}{*}{\([-1,1] \times [-1,1]\)}
& \multirow{2}{*}{$0.2$}
  & \deepgname & 98.0  & 460.93 &\\
&&& \methodname  & \textbf{98.0} & 1.34  & \\
\bottomrule
\end{tabular}
\caption{Comparison of \deepgname{} and \methodname{} on PGD networks with equivalent parameters (see Table~\ref{tab:experiment-params} Appendix~\ref{appendix:hyperparam}).
$^*$Results obtained with $P=10$ samples.  Interval size refers to the range split; time refers to the total certification duration.}
\label{tab:compare-deepG-superdeepg}
\end{table*}
\paragraph{Implementation}
\methodname{} is an open-source implementation integrated within the \texttt{auto\_LiRPA}~\cite{xu2020automatic} framework. The computation of the abstract domain $\mathcal{C}$ (described in section \ref{sec:fromdeepgtosuperdeepg})
is encapsulated in a \emph{layer} of a special type called \texttt{CustomOp}. 
To verify the network’s robustness (as in Equation \eqref{eq:gpropertygeom}), users adds this layer as the first layer to the neural network.
This simplifies the verification process. The verification of the neural network is done with the CROWN or CROWN-IBP method.
\paragraph{Baseline} We benchmark our method against state-of-the-art verifiers: \deepgname, FGV, and PWL. For PWL, although its code is available, a dependency on a MILP verifier prevented us from running it, so our comparison uses published results. We excluded Semantify-NN, as it has been proven both less precise than \deepgname and slower than FGV \cite{CGT}.
\paragraph{Experimental Setup}
FGV and \methodname experiments were conducted on an NVIDIA Tesla V100 PCIe (32 GB), while \deepgname, lacking GPU support, was run on an Intel Xeon E5-2660 v3 @ 2.60 GHz with 62 GB of RAM and 20 cores.
The evaluation is done on three key datasets used in prior robustness studies~\cite{deepG,CGT}.
We use the neural networks provided by \deepgname \cite{deepG}, referred to as \textit{PGD} and by \textit{CGT}~\cite{CGT}, referred to as \textit{CGT}. While the PGD and CGT networks use the same architecture, their weights differ as a result of different training strategies (see their respective papers for training details). Detailed descriptions of network architectures 
are provided in Appendix \ref{apx:archis-training}. 
To ensure our results are directly comparable to prior work, we adopt the evaluation protocols from the original papers: we evaluate the PGD networks on the first 100 test images, and the CGT networks on the first 10000 test images. For TinyImageNet, we use the pre-trained model \cite{vnncomp2024_tinyimagenet} from the VNN-Comp'24 benchmark \cite{brix2024fifth}, which was not specialized with geometric-robust training. We evaluate this model on the first 100 test images.
\paragraph{Metrics} 
We use the same metrics as most of the literature on neural network robustness verification~\cite{deepG, CGT}, namely the percentage of test samples guaranteed to be correctly classified under the specified transformations denoted as \textit{Certified (\%)}, and the certification time per image. For reference, we compute the \textit{Clean accuracy (\%)}, that refers to the percentage of test samples correctly classified without any transformation applied. We evaluate it on the same set of images used in our experiments, as it provides an upper bound for the \textit{Certified (\%)} metric. For each network architecture and each verifier, we evaluate one range of transformation parameters. Each range is split into fixed-size intervals. For example, a rotation range of $[-10^\circ, 10^\circ]$,  denoted R($10^\circ$), with a $2^\circ$ interval size (IS) is divided into: $[-10^\circ, -8^\circ], [-8^\circ, -6^\circ], \dots, [8^\circ, 10^\circ]$. 
\paragraph{Comparison with other methods}
First, we seek to determine whether \methodname{}
matches \deepgname{} in terms of certification rates ensuring that our reformulation performs similarly.
Second, we want to reach competitive verification times 
with respect to the faster tool FGV. Third, we aim to demonstrate that our method outperforms PWL.  The experimental validation is performed in two stages. First, to ensure our reformulation performs similarly, we compare \deepgname and \methodname using identical evaluation parameters.  The results are detailed in Table~\ref{tab:compare-deepG-superdeepg}. Note that \deepgname could not be run on the CGT model for portability issues.  Second, we adjust the interval size for each tool to compare runtimes at similar certification rates against FGV and PWL (see results in Tables  \ref{tab:compare-fgv-superdeepg} and \ref{tab:CIFAR_r10_sdg} for FGV and Observation 4 for PWL). 
These results lead to five key observations.
\setlength{\intextsep}{0pt}   
\setlength{\textfloatsep}{0pt}
\paragraph{Observation 1: \methodname is faster than \deepgname, with similar (or better) accuracy.} 
To compare with \deepgname, we use the smallest value of $N$ that satisfies Equation~\eqref{eq:linkepsilonN} which relates the number of subdivisions to the precision hyperparameter $\varepsilon$. We also use a Lipschitz upper-bound that corresponds to the maximum value $\mathcal L_{\max}$ observed across all pixels in the test images. This setup guarantees that our certified accuracy will be at least as high as that of \deepgname, which is confirmed by the experiments. Note that our CPU times are slower than those reported in the original paper (this is due to the use of a docker and a less performing CPU).
In Table~\ref{tab:compare-deepG-superdeepg}, 
\deepgname takes $210.72$s per MNIST image at \([-30^\circ,30^\circ]\) to certify $98$\%, whereas \methodname achieves $98$\% in 0.40s/image, up to \textbf{526×} faster. This trend persists for CIFAR10 (\([-10^\circ,10^\circ]\)) with interval size of $1^\circ$, 
where \deepgname takes $2389.70$s/image to certify 65\% and \methodname only 7.58s/image with 65\% certified images, making it approximately \textbf{315$\times$} faster. 
Overall, our method consistently accelerates the computation time.
\setlength{\intextsep}{0pt}   
\setlength{\textfloatsep}{0pt}
\paragraph{Observation 2: On CGT networks we match FGV's certified accuracy. } 
Table \ref{tab:compare-fgv-superdeepg} compares \methodname\ with FGV on CGT networks trained for FGV. To ensure fairness, we matched certified accuracy and compared runtimes (we adjust the interval size used by \methodname so that its certified accuracy matches).  On MNIST, both methods certify ~94\%, but FGV is significantly faster. On TinyImageNet, the runtime of \methodname remains within a small factor of FGV. On CIFAR10, both reach ~63\% accuracy, and \methodname\ is slightly faster, even though FGV is the fastest known method for geometric certification. This shows that while FGV performs best on networks tailored to it, \methodname\ remains competitive in both certified accuracy and speed, while not relying on any specialized geometric training.
\begin{table}[hbt]
\centering
\small
\begin{tabular}{lcccc} 
\toprule
\textbf{\makecell{Dataset \\ Transf. Clean acc.}}
& \makecell{\textbf{Interval} \\ \textbf{size} }
& \textbf{\makecell{Certif. \\ Method}}
& \textbf{\makecell{Cert. \\ Acc. (\%)}}
& \textbf{\makecell{Time \\ (s/im)}}
\\ 
\midrule 
  \multirow{2}{*}{\makecell{MNIST  \\  R(30\textdegree) 99.17\%}}
 & $0.25^\circ$
 & FGV      &  94.2  &  0.0020  \\
 &  $2^\circ$ & Ours   & 94.3 &  0.17 \\
\midrule
  \multirow{2}{*}{\makecell{CIFAR10 \\  R(10\textdegree) 80.47\%}}
 & $0.0002^\circ$
 & FGV      & 63.2 & 1.04 \\
 & $0.0125^\circ$ & Ours   & 63.1 & 0.97 \\
 \midrule
   \multirow{2}{*}{\makecell{TinyImageNet \\  Sh(2) 27.28\%}}
 & $0.00002$
 & FGV      & 18.7 & 0.17 \\
 & $0.001$ & Ours   & 19.4 & 0.54 \\
\midrule
  \multirow{2}{*}{\makecell{TinyImageNet \\  Sc(2) 26.12\%}}
 & $0.00002$
 & FGV      & 15.2 & 0.16 \\
  & $0.001$ & Ours   &  15.9 & 0.48  \\
\midrule
  \multirow{2}{*}{\makecell{TinyImageNet  \\  R(5\textdegree) 26.01\%}}
 & $0.001^\circ$
 & FGV      & 13.1  &  0.79  \\
 &  $0.08^\circ$ & Ours   & 13.0 & 1.28 \\
\midrule
\end{tabular}
\caption{\methodname performs effectively on models trained for the FGV certification method (CGT network).}
\label{tab:compare-fgv-superdeepg}
\end{table}
\setlength{\intextsep}{0pt}   
\setlength{\textfloatsep}{0pt}
\paragraph{Observation 3: On networks without geometric training, we achieve better precision and better runtime than FGV} 
On networks not trained with CGT, FGV is slower and overall less accurate than \methodname. Even with very small interval sizes, FGV still fails to match \methodname certification levels. For example, for a rotation in [$-10^{\circ}, 10^{\circ}$] on CIFAR10, FGV has a certification rate plateauing at 23$\%$ while \methodname reaches 65$\%$ in 1.90s/image. 
A smaller interval size leads to slower performance for FGV.
This is highlighted in Table~\ref{tab:CIFAR_r10_sdg} (highlighted column is the observed plateau). This trend is observed for other parameters and transformations (see Appendix \ref{apprendix:FGV_on_PGD}).
\setlength{\intextsep}{0pt}   
\setlength{\textfloatsep}{0pt}
\begin{table}[ht]
  \centering
  \small
  \begin{tabular}{lcc>{\columncolor[gray]{0.85}}cc c}
    \toprule
     R(10$^{\circ}$) & \multicolumn{4}{c}{FGV}   & 
    Ours \\
    \cmidrule(lr){2-5}\cmidrule(l){6-6}
      \makecell{Interval \\ size $({}^\circ)$} & $2\cdot10^{-3}$ & $2\cdot10^{-4}$ & $2\cdot10^{-5}$ & $2\cdot10^{-6}$ & $1$ \\
    \midrule
    \makecell{Certified \\ (\%)} & 6.0 & 19.0 & 23.0 & 23.0 & \textbf{65.0} \\
    Time (s)       & 0.57 & 5.39 & 44.40 & 443.00  & 1.90 \\
    \bottomrule
  \end{tabular}
  \caption{FGV plateaus on PGD networks, while \methodname achieves significantly better certified accuracy.}
  \label{tab:CIFAR_r10_sdg}
\end{table}
\setlength{\intextsep}{0pt}   
\setlength{\textfloatsep}{0pt}
\paragraph{Observation 4: Favorable comparison with PWL} PWL is designed for high precision over a few wide sub-intervals, whereas our method leverages GPU parallelization to efficiently process many sub-intervals. As demonstrated in Table~\ref{tab:pwl_comparison}, when we run our method under PWL's configuration (Interval size=6, P=1000, N=14500). PWL is more precise: $91\%$ certified accuracy versus our $86\%$ (aligning with the $92.9\%$ vs $87.8\% $ reported on 98 clean images in their paper). However, even in this non-optimal setting, our method is already faster (1.14s/image vs 28.3s/image). Our method's advantage appears when we use finer splits. By splitting the domain into more sub-intervals (Interval size=1), our accuracy increases to $98\%$. This approach also lets us reduce other hyper-parameters ($N$ from 14500 to 250) while maintaining $98\%$ accuracy. Using configuration (Interval size=1, P=10, N=250), our method achieves $98\%$ certified accuracy in $0.40$s/image, thereby outperforming the reported PWL configuration in both accuracy and runtime.
\begin{table}[hbt]
\centering
\small
\begin{tabular}{lccc}
\toprule
Tool & \makecell{Config \\(IS/P/N)}  & \makecell{Cert. (\%) \\  (98 clean im)} & \makecell{Time \\(s/im)} \\
\midrule
PWL           & 6 / 1000 / 14500 &  91.0 (92.9\%=91/98)  & 28.30     \\
Ours          & 6 / 1000 / 14500 &  86.0 (87.8\%=86/98)  & 1.14     \\
Ours          & 1 / 1000 / 14500 &  98.0 (100.0\%=98/98) & 8.54     \\
Ours & 1 / 10 / 250   & \textbf{98.0} (100.0\%=98/98)  & \textbf{0.40} \\
\bottomrule
\end{tabular}
\caption{Comparison with PWL using different configuration hyperparameters. Rotation in $[-30^\circ, 30^\circ]$ on MNIST.}
\label{tab:pwl_comparison}
\end{table}
\setlength{\intextsep}{0pt}   
\setlength{\textfloatsep}{0pt}
\paragraph{Observation 5: \methodname achieves significant geometric certification results on TinyImageNet without training specifically for geometric robustness.}
While \deepgname and PWL does not scale to TinyImageNet, \methodname does. On models not trained for FGV, \methodname achieves higher certified accuracy in significantly less time than FGV. For a [$-5^{\circ}, 5^{\circ}$] rotation, our method certifies 18$\%$ of images in 3.73s/image, while FGV runs out of memory (OoM) attempting. This trend is observed for other transformations (see Appendix \ref{apprendix:FGV_on_PGD}).  
\begin{table}[ht]
  \centering
  \small
  \setlength{\tabcolsep}{3pt} 
  \begin{tabular}{l ccc>{\columncolor[gray]{0.85}}c cc }
    \toprule
    R(5$^{\circ}$)  & \multicolumn{4}{c}{FGV} & \multicolumn{2}{c}{Ours } \\
    \cmidrule(lr){2-5}\cmidrule(l){6-7}
    Interval size $({}^\circ)$ & $4\cdot10^{-4}$ & $3\cdot10^{-4}$ & $2\cdot10^{-4}$ & $1\cdot10^{-4}$ & 0.04 & 0.02 \\
    \midrule
    Certified (\%) & 2.0 & 2.0 & 10.0 & OoM & 10.0 & \textbf{18.0} \\
    Time (s)       & 5.06 & 6.78 & 10.90 & --  & 1.46 & \textbf{3.73} \\
    \bottomrule
  \end{tabular}
  \caption{On TinyImageNet VNN-Comp'24 model, \methodname achieves higher accuracy than FGV in less time. }
  \label{tab:tinyImageNet}
\end{table}
\paragraph{Influence of the hyper-parameters P and interval size.} 
Step 1(-bis) could be considered useless, since Step 3 can correct any inaccurate bounds. Experiments still show that performing Step 1(-bis)
does significantly improve the precision.
However, computing too many sample points is unnecessary as shown in Figure~\ref{fig:nb_samples_step1}. We use a large interval ($\text{Interval size}=6^\circ$) to reveal potential variation, but observe no major improvement beyond $P=10$. For small intervals ($\text{Interval size}=0.5^\circ$), certified accuracy is constant from the start.
\begin{figure}[hbt]
    \centering
\includegraphics[width=0.80\linewidth]{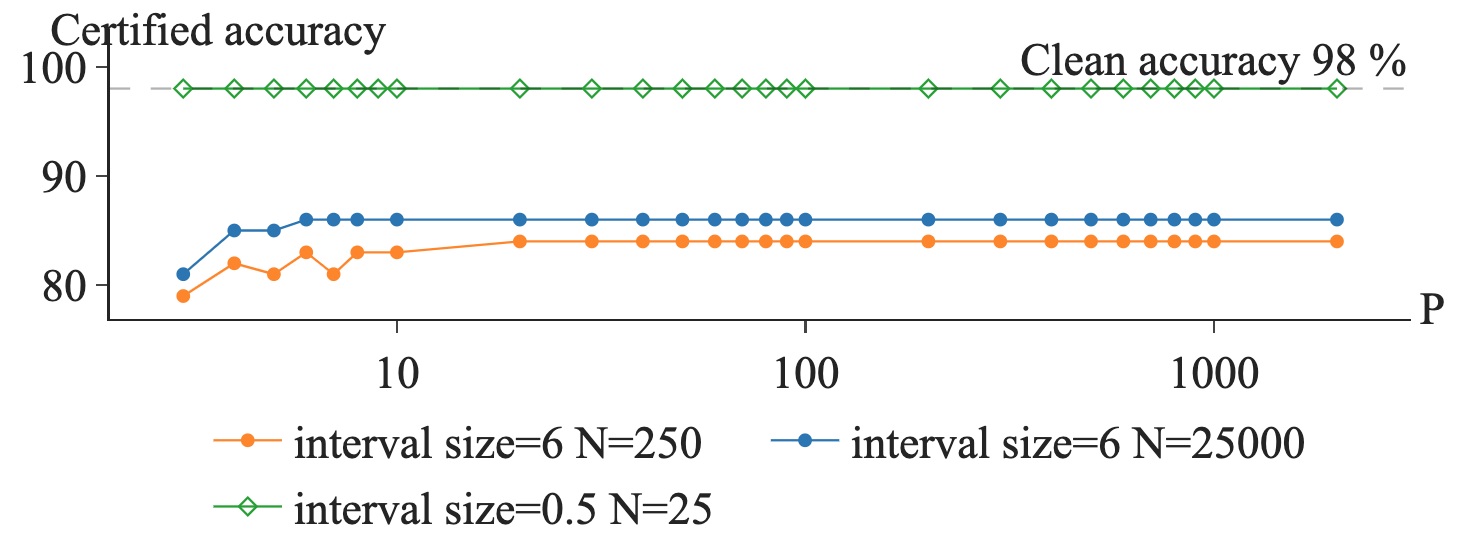}
    \caption{Increasing $P$ beyond 10 brings no significant gains (log scale). Rotation in $[-30^\circ, 30^\circ]$ on MNIST. } 
    \label{fig:nb_samples_step1}
\end{figure}
\paragraph{Influence of the hyper-parameters N and interval size} 
Reducing the interval size improves certified accuracy significantly, reaching the clean accuracy baseline, as illustrated in Figure \ref{fig:interval_size_influence} with $P{=}10$ and rotations in $[-30^\circ, 30^\circ]$. Increasing $N$ from 250 to 25000 yields marginal benefits while increasing computation time. The optimal trade-off is reached at an interval size of 1, where $N{=}250$ gives the optimal certified accuracy with 0.40\,s/image. 
\begin{figure}[hbt]
    \centering
\includegraphics[width=0.55\linewidth]{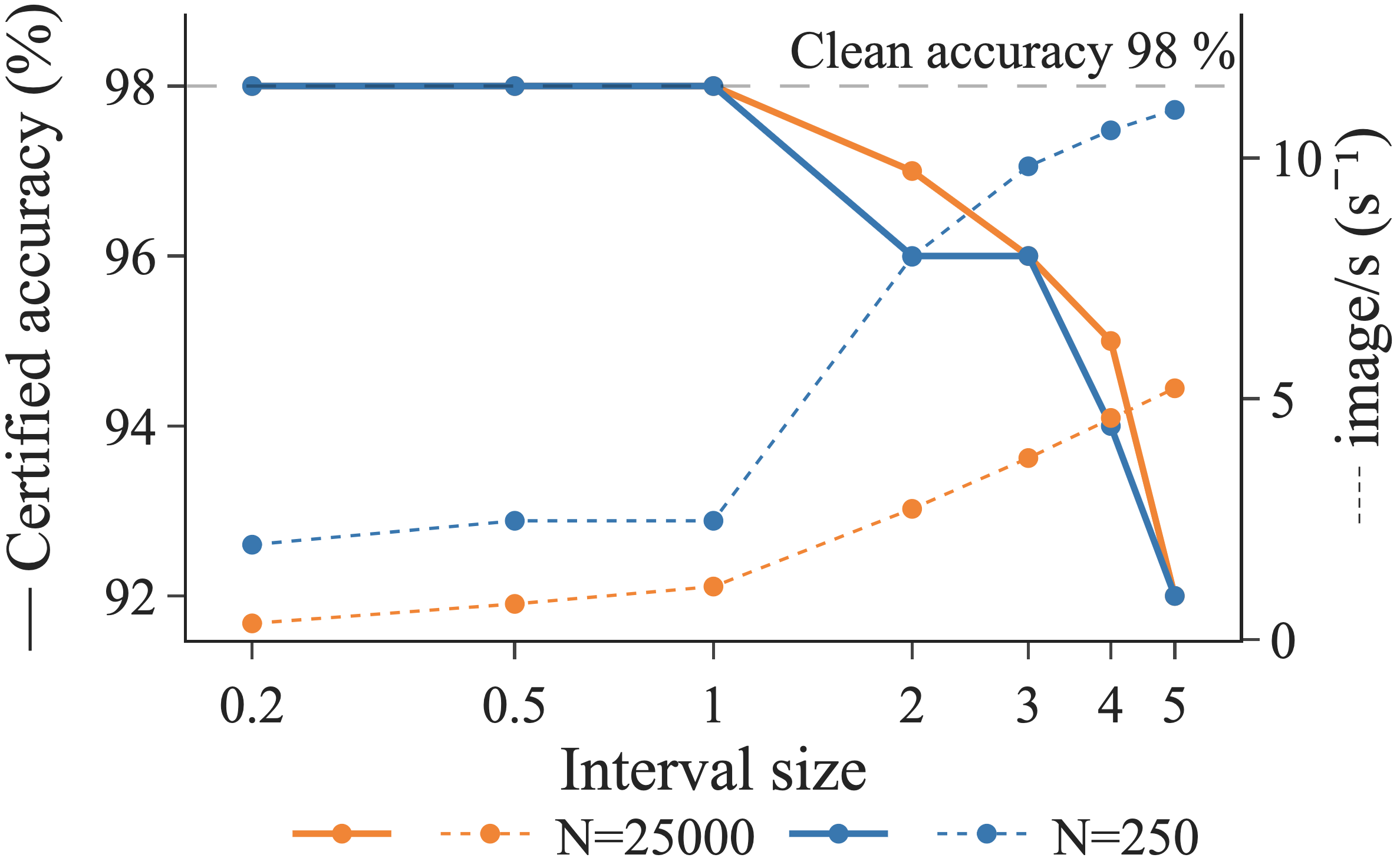}
    \caption{Smaller interval sizes improve certified accuracy more than increasing $N$ (rotations in $[-30^\circ, 30^\circ]$ on MNIST shown), log scale. } 
    \label{fig:interval_size_influence}
\end{figure}

\section{Conclusion}\label{sec:conclusion}

We proposed \methodname a method for computing linear constraints on geometric
image transformations. Our work builds upon \deepgname, which uses Lipschitz optimization to compute tight linear relaxations. However, both \deepgname and subsequent precision-focused methods like PWL rely on computationally expensive solvers that limit their scalability. Inspired by FGV, which achieve remarkable speed via GPU-optimization and simpler interval-based abstractions, we pursued a new direction. Instead of sacrificing precision for speed, we reformulated the core optimization problems of \deepgname to be analytically solvable and fully \textit{tensorized}. This redesign eliminates computational bottlenecks and enables massive GPU parallelization. Although this work currently focuses on classification, the next step for runway detection is to extend our method to object detection and its specific constraints.
Future work will also extend this framework to certify a broader set of transformations and more complex safety properties.



\appendix

\section{Reminder of geometric transformations on image coordinates}\label{appendix:geom_transfo}
The inverse formulas of the geometric transformations used in this paper applied to an image coordinate $(x,y)$ are:
\begin{table}[hbt]
\centering
\begin{tabular}{cc}
\toprule
Rotation $R_\theta^{-1}(x, y)$ & Translation $T_{v_{1},v_{2}}^{-1} (x,y)$ \\
\toprule
$\displaystyle
    \begin{pmatrix}
        \cos\theta & \sin\theta \\
        -\sin\theta & \cos\theta
    \end{pmatrix}
    \begin{pmatrix}
        x \\
        y
    \end{pmatrix}
$
& 
$\displaystyle
     \begin{pmatrix} x - v_{1} \\ y - v_{2} \end{pmatrix}
$
\\ \addlinespace[0.5em] 
\toprule
Scaling $Sc_{\lambda}^{-1} (x,y)$  & Shearing $Sh_{m}^{-1} (x,y)$\\
\toprule
$\displaystyle
    \begin{pmatrix} \frac{1}{ \lambda} & 0\\ 0&\frac{1}{\lambda}  \end{pmatrix}
    \begin{pmatrix} x\\ y  \end{pmatrix}
$
& 
$\displaystyle
    \begin{pmatrix} 1 & -m \\ 0 &1 \end{pmatrix}
    \begin{pmatrix} x\\ y  \end{pmatrix}
$
\\
\bottomrule
\end{tabular}
\end{table}
\section{Supplementary details for \methodname steps}
\subsection{Proof by contradiction}\label{proofbycontradiction}
Let us consider problem \eqref{eq:LP}, and let us assume that it accepts an optimal solution \((\underline{A}^*, \underline{B}^*)\) that passes by none of the points $\big(\theta^p, g_x(\theta^p)\big)$ for $p = 1, \ldots, P$. This means that all residuals are strictly positive. Let $\delta$ denote the (strictly positive) value of the smallest residual: $\delta := \min_{p = 1,\ldots, P} \underline{r}(\theta^p; \underline{A}^*, \underline{B}^*)$. We have: $$ \underline{r}(\theta^p; \underline{A}^*, \underline{B}^*) \geq \delta > 0, \quad \text{for } p=1,\ldots,P$$. Let us consider the candidate obtained by shifting the solution by $\delta/2$: $$\underline{A}' = \underline{A}^* \qquad \underline{B}' = \underline{B}^* + \delta/2$$. All the residuals remain positive for $(\underline{A}', \underline{B}')$, which means it satisfies the constraints:
\begin{align*}
\underline{r}(\theta^p; \underline{A}', \underline{B}')
&=g_x(\theta^p) - \underline{A}'\cdot\theta^p - \underline{B}'\\
&=g_x(\theta^p) - \underline{A}^*\cdot\theta^p - \underline{B}^* - \delta/2\\ 
&= \underline{r}(\theta^p; \underline{A}^*, \underline{B}^*) - \delta/2 \\
&\geq \delta - \delta/2 = \delta/2 
\end{align*}
The objective value of $(\underline{A}', \underline{B}')$ is less than that of \((\underline{A}^*, \underline{B}^*)\):
\begin{align*}
\frac{1}{P} \sum_{p=1}^{P} \left[ \underline{r}(\theta^p;\underline{A}^\prime,\underline{B}^\prime)\right]
&= \frac{1}{P} \sum_{p=1}^{P} \underline{r}(\theta^p; \underline{A}^*, \underline{B}^*) -\delta/2
\end{align*}
This contradicts our assumption that \((\underline{A}^*, \underline{B}^*)\) is optimal. This proves by contradiction that if \((\underline{A}^*, \underline{B}^*)\) is an optimal solution, there exists at least one \( p^* \) such that $\underline{r}(\theta^{p^*}; \underline{A}^*, \underline{B}^*) = 0$. \qed
\subsection{Step 1-bis and 3-bis for multi-dimensional inputs}\label{appendix:multi-D}
The two following sections 
extend the mathematical formulations to the general case where $\kappa \in \mathbb{R}^d$. The main paper focused on the one-dimensional case $\kappa = \theta \in \mathbb{R}$, corresponding to transformations such as rotation, scaling, or shearing. Here, we provide details for higher-dimensional transformations, including translation ($d=2$).
\subsection{ Step 1-bis: Analytically unsound bounds for multi-dimensional inputs}\label{sec:multidunsound}
Let $\kappa\in\mathbb{R}^{d}$ be a perturbation vector bounded by
$\underline{\kappa}\le \kappa\le\overline{\kappa}$ such that $ [\underline\kappa,\overline\kappa]= [\underline\kappa_1,\overline\kappa_1]\times
[\underline\kappa_2,\overline\kappa_2]\times\cdots\times
[\underline\kappa_d,\overline\kappa_d]$.
The geometric transformation $g_x:\mathbb{R}^{d}\to\mathbb{R}^{n \times m}$, obtained by composing the $d$-dimensional geometric transformations
with an interpolation, associates an image to any
parameter vector $\kappa$.
We sample \(P\) affinely independent vectors
$\kappa^{(p)} = (\kappa_1^{(p)}, \ldots, \kappa_d^{(p)}) \quad p = 1, \dots, P$,
from the domain $[\underline{\kappa}, \overline{\kappa}]$ and compute the P associated images \(g_x(\kappa^{(p)})\).
The objective is to find the tightest pair of hyperplanes,
a lower one with coefficients $(\underline{A},\underline{B}) \in (\mathbb{R}^{d \times  n\times m}, \mathbb{R}^{n\times m})$
and an upper one with coefficients $(\overline{A},\overline{B}) \in (\mathbb{R}^{d \times n\times m}, \mathbb{R}^{n\times m})$,
that encloses every sampled point $(\kappa^{(p)},\,g_x(\kappa^{(p)}))$, i.e,
 $\forall \kappa \in \{\kappa^{(1)}, ..., \kappa^{(P)}\} $: 
\begin{equation}
\sum_{i=1}^d \kappa_i\underline{A}_i\, + \underline{B} \leq g_x(\kappa) \leq \sum_{i=1}^d \kappa_i\overline{A}_i\, + \overline{B}
\label{eq:samples_encloses_gx}
\end{equation}
To proceed, introduce the residual functions:
\begin{equation*}
    \underline{r}(\kappa;\underline{A},\underline{B}) =  g_x(\kappa) - \sum_{i=1}^d \kappa_i\underline{A}_i\, + \underline{B}
\end{equation*}
\begin{equation*}
\overline{r}(\kappa;\overline{A},\overline{B}) =  g_x(\kappa) - \sum_{i=1}^d \kappa_i\overline{A}_i\, + \overline{B}
\end{equation*}
For the sampled points, the constraints from~\eqref{eq:samples_encloses_gx} become:
\begin{equation*}
    \underline{r}(\kappa;\underline{A},\underline{B}) \geq 0 , \quad \overline{r}(\kappa;\overline{A},\overline{B}) \leq 0 \qquad \forall \kappa \in \{\kappa^{(1)}, ..., \kappa^{(P)}\}
\end{equation*}
To minimize the gap between the affine bound and the sampled values, the average residual should be as small as possible. Therefore, $\underline{A}, \underline{B}$ are the solutions of an optimization problem that minimizes the average residual while enforcing negativity for every sample point. A similar process is applied to compute an upper bound. This is formulated as the following \(\mathfrak{P}\) optimization problem:
\begin{equation}
\begin{aligned}
\min_{\underline{A}, \underline{B}} \quad & \left\{ \frac{1}{P} \sum_{p=1}^{P} \underline{r}(\kappa^{(p)}; \underline{A}, \underline{B}) \right\} \\
\text{subject to} \quad & \underline{r}(\kappa^{(p)}; \underline{A}, \underline{B}) \geq 0, \quad \forall p = 1,\ldots, P
\end{aligned}    
\tag{$\mathfrak{P}$}
\end{equation}
\textbf{Proposition 1.}
The optimization problem $\mathfrak{P}$ accepts an optimal solution $(\underline A,\underline B)$ such that there exists a subset of $d+1$ affinely independent sample indices
\(\mathcal J=\{p_1,\dots,p_{d+1}\}\subset\{1,\dots,P\}\)
that verify: 
\begin{equation}
    \sum_{i=1}^d \kappa_i^{(p)} \underline{A}_i\, + \underline{B} = g_x\bigl(\kappa^{(p)}\bigr),
\quad \forall p \in \mathcal J
\label{eq:residual_0}
\end{equation}
\textbf{Proof of proposition 1.} We consider the optimization problem $\mathfrak{P}$ for a fixed pixel position \((u,v)\), where the affine coefficients are denoted by \((\underline{a}, \underline{b}) \in \mathbb{R}^d \times \mathbb{R}\). The residual constraint for sample \(p\) can be written as: $\kappa^{(p)\top} \underline{a} + \underline{b} \leq p_{u,v}(\kappa^{(p)}), \quad \forall p = 1, \dots, P$.
Define the stacked variable \(z := (\underline{a}^\top, \underline{b})^\top \in \mathbb{R}^{d+1}\). Then the constraints define a polyhedron of the form: $\{z \in \mathbb{R}^{d+1} \mid Cz \leq e\}$
where each row of \(C \in \mathbb{R}^{P \times (d+1)}\) is given by \((\kappa^{(p)\top}, 1)\), and \(e \in \mathbb{R}^P\) is such \(e_p = p_{u,v}(\kappa^{(p)})\). Any set of \(d+1\) linearly independent active constraints defines a unique basic solution of this polyhedron, at least one of which is optimal \cite{hillier2015introduction, bertsimas1997introduction, StanfordScanRep}. These correspond to a subset of \(d+1\) affinely independent sample points \(\kappa^{(p)}\) for which the inequality holds with equality: $\kappa^{(p)\top} \underline{a} + \underline{b} = p_{u,v}(\kappa^{(p)}), \quad \forall p \in \mathcal{J},$ for some index set \(\mathcal J=\{p_1,\dots,p_{d+1}\}\subset\{1,\dots,P\}\), with \(|\mathcal{J}| = d+1\) and the \(\{\kappa^{(p)}\}_{p \in \mathcal{J}}\) affinely independent. Applying the same reasoning to each pixel position \((u,v)\), we obtain equation~\eqref{eq:residual_0}.
Based on proposition 1, we solve $\mathfrak{P}$ by enumerating all $\binom{P}{d+1}$ subsets of $d+1$ affinely independent samples, computing the corresponding $(\underline{A}, \underline{B})$, verifying feasibility with respect to the residual constraints, and retaining the feasible candidate that yields the smallest objective value.
\subsubsection{Select $d+1$ distinct parameters}\label{subsec:first}
In light of Proposition~1, we select a subset of $d+1$ affinely independent parameter vectors \(\kappa^{(1)},\dots,\kappa^{(d+1)} \in \mathbb{R}^d\). We assume that the optimal affine lower bound passes through these points, i.e., they satisfy Equation~\eqref{eq:residual_0}.
\subsubsection{Eliminate $\underline{B}$}
 Subtract the equation at $p=d+1$ from Equation \eqref{eq:residual_0} for $p=1,\dots,d$:
   \[
     \sum_{i=1}^d \,(\kappa_i^{(p)} - \kappa_i^{(d+1)})\underline{A}_i
     = g_x\bigl(\kappa^{(p)}\bigr) - g_x\bigl(\kappa^{(d+1)}\bigr) 
   \]
\subsubsection{Solving the $d\times d$ system}
Collect the $d$ equations obtained above into matrix form.
Write $p = 1,\dots ,d\;\;\;\;\; i = 1,\dots ,d$:
\[
   K_{p i} := (\kappa_i^{(p)} - \kappa_i^{(d+1)}) \in \mathbb{R}^{d\times d}
\]
\[
   G_{p}    := g_x\bigl(\kappa^{(p)}\bigr) - g_x\bigl(\kappa^{(d+1)}\bigr) \in \mathbb{R}^{d\times n \times m}
\]
The problem reduces to a linear system where the unknown is the slope tensor $\underline{A} \in \mathbb{R}^{d \times n\times m }$: $K \underline{A} = G$.
Because the $d+1$ selected points are affinely independent, the matrix $K$ has full rank and is therefore invertible. We obtain a unique vector: $\underline A=K^{-1} G$.
With $\underline A$ known, the bias is recovered from any
 point, for $k=1$:
  $\underline B=
  g_x\!\bigl(\kappa^{(1)}\bigr)-
  \sum_{i=1}^{n}\underline A_i\,\kappa^{(1)}_i $
\subsubsection{Feasible candidates}\label{subsec:last}
If $\exists p \in 1,\ldots, P$ such that $\underline{r}(\kappa^{(p)}; \underline{A}, \underline{B}) \leq 0 $.
The chosen candidate must be discarded.
\subsubsection{Enumerating candidates}
We repeat this procedure for all subsets of $d+1$ points among the $P$ possible parameter combinations.  
For each subset, we retain the pair $(\underline A, \underline B)$, and finally select the one that yields the lowest value for the optimisation problem~\eqref{eq:LP}. A similar process is applied for the upper bound $(\overline A, \overline B)$. In practice, this enumeration is not a computational bottleneck, as experiments confirm that a very small number of sample points P is sufficient (see Figure \ref{fig:nb_samples_step1}), while the dimension d of the geometric transformations is low.
\subsection{Step 3-bis: Correction of the approximation error for multi-dimensional inputs }\label{sec:multidsound}
The bounds computed in the previous step hold for the sampled points, but they may fail to bound $g_x(\kappa)$ from below over the full domain $[\underline{\kappa}, \overline{\kappa}]$. 
The goal of this step is to add a correction term $\delta^{*}$ in order to have a valid lower bound over the entire domain:
\begin{equation*}
    \forall \kappa \in [\underline{\kappa}, \overline{\kappa}] \;\;\;\;\;\;\;\;\; \;\;\;\;\;\;\sum_{i=1}^d \kappa_i \underline{A}_i + \underline{B} - g_x(\kappa) + \underline{\delta}^* \leq 0
\end{equation*}
where $\underline{\delta}^* =\min\limits_{\kappa \in [\underline{\kappa}, \overline{\kappa}]}\underline{r}(\kappa, \underline{A}, \underline{B})$. Hovewer, computing the exact minimum of $\underline{r}(\kappa, \underline{A}, \underline{B})$ is computationally hard due to the nonlinearity and nonconvexity of the function. To do so, we compute an approximation $\underline{\delta} \leq \underline{\delta}^*$ by partitioning the domain and applying the  Lipschitz continuity (with a Lipschitz constant per dimension) on each partition. The residual \(\underline r\) is evaluated over the hyper‑rectangle:
\[
[\underline\kappa,\overline\kappa] = [\underline\kappa_1,\overline\kappa_1]\times
[\underline\kappa_2,\overline\kappa_2]\times\cdots\times
[\underline\kappa_d,\overline\kappa_d]
\]
We subdivide each dimension interval into \(N\) equal parts, so that for each \(i=1,\dots,d\) we have: $[\underline\kappa_i,\overline\kappa_i]
=\bigcup_{\ell=0}^{N-1}\bigl[\kappa_i^{(\ell)},\,\kappa_i^{(\ell+1)}\bigr] $.
This defines the uniform spacing between consecutive subdivision points:
\begin{equation}
\Bigl\lvert \kappa_i^{(\ell+1)}-\kappa_i^{(\ell)} \Bigr\rvert\ =\frac{\overline\kappa_i-\underline\kappa_i}{N}
\label{eq:interval_size}
\end{equation}
Let us denote by \(c_i^{(\ell)}=(\kappa_i^{(\ell+1)}+\kappa_i^{(\ell)})/2 \in \mathbb{R}\) the center of one such partition.
The maximum distance between any point in a sub-interval and its center equals half the sub-interval width, using \eqref{eq:interval_size}:
\[
\max_{\kappa_i\in[\kappa_i^{(\ell)},\,\kappa_i^{(\ell+1)}]}
\Bigl\lvert\,
\kappa_i-c_i^{(\ell)}\Bigr\rvert
=\frac{\overline\kappa_i-\underline\kappa_i}{2N}
\]
Each of the \(d\) dimensions is divided into \(N\) equal parts, yielding \(N^d\) total cells. Let \(C^h\in\mathbb{R}^d\) denote the center of the \(h\)-th cell. Evaluating the residual at these centers, we define:
\[
m \;=\;\min_{h=0,\dots,N^d}\;
\underline r\bigl(C^{h};\,\underline A,\underline B\bigr)
\]
Finally, using Lipschitz continuity (with a Lipschitz constant per dimension), we obtain the correction term:
\[
m
\;-\;\sum_{i=1}^{d}\mathcal \max_{\kappa}\bigl|\partial_i \underline{r}(\kappa)\bigr|\,
\frac{\overline\kappa_i-\underline\kappa_i}{2 N}
\;\leq\;
\min\limits_{\kappa \in [\underline{\kappa}, \overline{\kappa}]}\underline{r}(\kappa, \underline{A}, \underline{B})
\]
And we use this bound as a correction term:
\[
\underline{\delta} = m
\;-\;\sum_{i=1}^{d}\mathcal \max_{\kappa}\bigl|\partial_i \underline{r}(\kappa)\bigr|\,
\frac{\overline\kappa_i-\underline\kappa_i}{2N}
\]
The same process is applied for the upper bound. Finally, we have:
\begin{equation*}
\forall \kappa \in [\underline{\kappa}, \overline{\kappa}], \; \; \sum_{i=1}^d \kappa_i \underline{A}_i\, + \underline{B} +\underline{\delta} \leq g_x(\kappa) \leq \sum_{i=1}^d \kappa_i \underline{A}_i + \underline{B}+\overline{\delta}
\end{equation*}
\section{Supplementary details for the experiments section}
\subsection{Neural network architectures}\label{apx:archis-training}
We describe the network architectures for each dataset below, representing a convolutional layer as a 4-tuple consisting of (number of filters, kernel size, stride, and padding) as presented in \cite{CGT}:
\begin{itemize}
    \item \textbf{MNIST:} 2 conv layers $\{(32, 4, 2, 1), (64, 4, 2, 1)\}$ followed by 2 linear layers with $\{200, 10\}$ neurons. All layers are followed by a ReLU activation, except for the final output layer.
    \item \textbf{CIFAR10:} 3 conv layers $\{(32, 3, 1, 1)$, $(32, 4, 2, 1)$, $(64, 4, 2, 1)\}$ followed by 2 linear layers with $\{150, 10\}$ neurons. All layers are followed by a ReLU activation, except for the final output layer.
    \item \textbf{TinyImageNet:} 5 conv layers $\{(64, 3, $1$, 1)$, $(64, 3, 1, 1)$, $(128, 3, 2, 1)$, $(128$, $3$, $1$, $1)$, $(128, 3, 2, 1)\}$ followed by 2 linear layers with $\{512, 200\}$ neurons. Each conv layer is followed by a batch norm layer then a ReLU activation. The first linear layer is followed by a ReLU activation.
\end{itemize}
For the architectures described above, the network weights are sourced from the PGD-trained models presented in \cite{deepG} and the CGT-trained models from \cite{CGT}.
\subsection{Table of hyper-parameters linking \deepgname and \methodname}\label{appendix:hyperparam}
Table~\ref{tab:experiment-params} lists the hyper-parameter configurations used to align \methodname\ with \deepgname\ for fair comparison.
\begingroup
\begin{table}[hbt]
\centering
\begin{tabular}{lrrrrr}
\toprule
    N° & \textbf{$\mathcal L_{\max}$} & \makecell{Interval \\ size} & \textbf{$\varepsilon$} & \textbf{$N_{\max}$} & P \\
\midrule
1A   & 29.00  & 1\textdegree  & 0.001  & 247 & 10        \\
1B   & 29.00  & 6\textdegree    & 0.0001 & 14500 & 1000        \\
2  & 22.20  & 1\textdegree  & 0.0001 & 1887   & 1000     \\
3            & 29.41  & 0.040  & 0.006  & 98    & 1000    \\
4           & 50.67  & 0.010  & 0.001  & 254     & 1000    \\
5               & 19.12  & 0.040  & 0.006  & 64   & 1000      \\
6              & 26.31  & 0.020  & 0.0001 & 2631  & 100     \\
\bottomrule
\end{tabular}
\caption{Parameter selection for comparison with \deepgname. Configuration numbers appear in the first column (Table~\ref{tab:compare-deepG-superdeepg}).}
\label{tab:experiment-params}
\end{table}
\endgroup
\subsection{ Parameter configurations for experiments}
The hyperparameter configurations for our experiments are available in \texttt{Experiment/specification}. In Table~2 we selected parameters equivalent to those of \deepgname, while the parameters in Table~3 were chosen to ensure identical certified accuracy for both FGV and \methodname.
\subsection{Comparison with FGV on networks without geometric training}\label{apprendix:FGV_on_PGD}
Table \ref{tab:FGV_plateaus} shows the certification rates of FGV and \methodname on PGD-trained and VNNComp’24 networks. For FGV, the percentage of certified images slightly increases as the interval size decreases, but we observe that it would not be interesting to reduce the interval size any further, since the calculation time grows rapidly and the certification percentage reaches a plateau. Meanwhile, \methodname achieves significantly better certified accuracy.  Here, R, Sc, Sh, and T stand for rotations, scaling, shearing, and translations, respectively. 
\begin{table}[hbt]
\centering
\resizebox{0.50\textwidth}{!}{%
\begin{tabular}{c c c c c}
\toprule
\textbf{\makecell{Transf. \\ Dataset \\ Clean acc.}}
& \textbf{Tool}
& \textbf{\makecell{Interval \\ size}}
& \textbf{\makecell{Cert. \\ (\%)}}
& \textbf{\makecell{Time \\ per im \\ (s)}}
\\ \midrule

\multirow{4}{*}{\makecell{R(10)\\CIFAR10\\PGD\\(74.0\%)}}  
  & \multirow{4}{*}{FGV}  
      & $2\times10^{-3}$ & 6.0   & 0.57 \\
  &                                            & $2\times10^{-4}$ & 19.0  & 5.39  \\
  &                                            & $2\times10^{-5}$ & 23.0  & 44.40  \\
  &                                            & $2\times10^{-6}$ & 23.0  & 443   \\
\cmidrule(lr){2-5}
\multirow{1}{*}{} & \multirow{1}{*}{ours} & $1$ & \textbf{65.0} & 1.90 \\
\midrule
\multirow{5}{*}{\makecell{Sc(1)\\CIFAR10\\PGD\\(79.0\%)}}  
  & \multirow{5}{*}{FGV}  
      & $5\times10^{-5}$ & 5.0   & 0.044 \\
  &                                            & $5\times10^{-6}$ & 19.0  & 0.23  \\
  &                                            & $5\times10^{-7}$ & 24.0  & 2.02   \\
  &                                            & $5\times10^{-8}$ & 24.0  & 19.70   \\
  &                                            & $5\times10^{-9}$ & 24.0  & 166    \\
\cmidrule(lr){2-5}
\multirow{1}{*}{} & \multirow{1}{*}{ours} & 0.01 & \textbf{75.0} & 0.20 \\
\midrule
\multirow{5}{*}{\makecell{Sh(2)\\CIFAR10\\PGD\\(72.0\%)}}  
  & \multirow{5}{*}{FGV}  
      & $2.5\times10^{-4}$ & 3.0   & 0.034 \\
  &                                            & $2.5\times10^{-5}$ & 20.0  & 0.10  \\
  &                                            & $2.5\times10^{-6}$ & 22.0  & 0.80  \\
  &                                            & $2.5\times10^{-7}$ & 23.0  & 7.54   \\
  &                                            & $2.5\times10^{-8}$ & 23.0  & 67.50   \\
\cmidrule(lr){2-5}
\multirow{1}{*}{} & \multirow{1}{*}{ours} & 0.02 & \textbf{69.0} & 0.18 \\
\midrule
\multirow{4}{*}{\makecell{R(30)\\MNIST\\PGD\\(98.0\%)}}  
  & \multirow{4}{*}{FGV}  
      & $2.5\times10^{-1}$ & 0.0   & 0.033 \\
  &                                            & $2.5\times10^{-2}$ & 58.0  & 0.30  \\
  &                                            & $2.5\times10^{-3}$ & 70.0  & 2.81   \\
  &                                            & $2.5\times10^{-4}$ & 70.0  & 28.00   \\
\cmidrule(lr){2-5}
\multirow{1}{*}{} & \multirow{1}{*}{ours} & $1$ & \textbf{98.0} & 0.40 \\
 \midrule
\multirow{5}{*}{\makecell{R(5)\\TinyImageNet\\VNNComp'24\\(59.0\%)}}
  & \multirow{5}{*}{FGV}
    & $5\times10^{-4}$ & 1.0     & 4.05  \\
  & & $4\times10^{-4}$ & 2.0     & 5.06  \\
  & & $3\times10^{-4}$ & 2.0     & 6.78  \\
  & & $2\times10^{-4}$ & 10.0    & 10.09 \\
  & & $1\times10^{-4}$ & /     & OOM   \\
\cmidrule(lr){2-5}
\multirow{1}{*}{} & \multirow{1}{*}{Ours} & $0.02$ & \textbf{18.0} & 3.73 \\
\midrule
\multirow{5}{*}{\makecell{Sc(2)\\TinyImageNet\\VNNComp'24\\(59.0\%)}}
  & \multirow{5}{*}{FGV}
    & $5\times10^{-6}$ & 7.0     & 1.80  \\
  & & $1\times10^{-6}$ & 30.0    & 9.08  \\
  & & $7\times10^{-7}$ & 35.0    & 12.96 \\
  & & $6\times10^{-7}$ & 35.0    & 24.04 \\
  & & $5\times10^{-7}$ & /     & OOM   \\
\cmidrule(lr){2-5}
\multirow{1}{*}{} & \multirow{1}{*}{Ours} & 0.0002 & \textbf{36.0} & 4.34 \\
\midrule
\multirow{5}{*}{\makecell{Sh(2)\\TinyImageNet \\VNNComp'24\\(59.0\%)}}
  & \multirow{5}{*}{FGV}
    & $3\times10^{-6}$ & 38.0    & 2.99  \\
  & & $2\times10^{-6}$ & 41.0    & 4.48  \\
  & & $1\times10^{-6}$ & 47.0    & 8.96  \\
  & & $9\times10^{-7}$ & 47.0    & 9.95  \\
  & & $8\times10^{-7}$ & 47.0    & 11.23 \\
\cmidrule(lr){2-5}
\multirow{1}{*}{} & \multirow{1}{*}{Ours} & 0.0003 & \textbf{48.0} & 3.16 \\
\bottomrule
\end{tabular}}
\caption{On networks without geometric training, \methodname achieves significantly better results than FGV.}
\label{tab:FGV_plateaus}
\end{table}

\bibliographystyle{abbrv}
\bibliography{sample-base}

\end{document}